\documentclass{article}

    \usepackage[preprint]{neurips_2020}

\usepackage{natbib}
\setcitestyle{square,numbers}
\usepackage[utf8]{inputenc} % allow utf-8 input
\usepackage[T1]{fontenc}    % use 8-bit T1 fonts
\usepackage{hyperref}       % hyperlinks
\usepackage{url}            % simple URL typesetting
\usepackage{booktabs}       % professional-quality tables
\usepackage{amsfonts}       % blackboard math symbols
\usepackage{nicefrac}       % compact symbols for 1/2, etc.
\usepackage{microtype}      % microtypography

\usepackage{enumitem}
\setlist[itemize]{noitemsep, nolistsep}
\usepackage{amsmath}
\usepackage{multirow}
\usepackage{graphicx}
\usepackage{caption}
\usepackage{subcaption}
\usepackage[symbol]{footmisc}

\makeatletter
\newcommand\footnoteref[1]{\protected@xdef\@thefnmark{\ref{#1}}\@footnotemark}
\makeatother

\title{Neural Text Classification by Jointly Learning to Cluster and Align}

\author{%
  Yekun Chai$^{1}$\thanks{Equal contribution.} \quad Haidong Zhang$^{1}$\footnotemark[1] \quad Shuo Jin$^{2}$\\
  $^{1}$Institute of Automation, Chinese Academy of Sciences\\
  $^{2}$University of Pittsburgh\\
 \texttt{chaiyekun@gmail.com} \quad \texttt{haidong\_zhang14@yahoo.com} \\
}

\begin{document}

\maketitle

\begin{abstract}
Distributional text clustering delivers semantically informative representations and captures the relevance between each word and semantic clustering centroids. We extend the neural text clustering approach to text classification tasks by inducing cluster centers via a latent variable model and interacting with distributional word embeddings, to enrich the representation of tokens and measure the relatedness between tokens and each learnable cluster centroid. The proposed method jointly learns word clustering centroids and clustering-token alignments, achieving the state of the art results on multiple benchmark datasets and proving that the proposed cluster-token alignment mechanism is indeed favorable to text classification. Notably, our qualitative analysis has conspicuously illustrated that text representations learned by the proposed model are in accord well with our intuition.
\end{abstract}

\section{Introduction}
\label{sec:intro}
% overview + widespread applications
Text classification, as an extensively applied fundamental cornerstone for natural language processing (NLP) applications, such as sentiment analysis~\cite{xue2018aspect}, spam detection~\cite{kennedy2019fact} and spoken dialogue systems~\cite{lowe2016evaluation,gupta2019simple}, has been widely studied for decades. In general, almost all NLP tasks can be cast into classification problems on either document, sentence, or word level. Here we are focusing on the means of it in a narrow sense, \emph{i.e.}, given a sequence of tokens with arbitrary length, predicting the most likely categorization it belongs to.

% conventional approaches, CNN/LSTM pros, + cons: lack the efficacy to capture the latent representations.
Considerable compelling neural approaches to the text classification task have empirically demonstrated their remarkable behaviors in recent years, to whom how to orchestrate and compose the semantic and syntactic representations from texts are central. Much of the work concentrated on learning the composition of distributional word representations~\cite{mikolov2013efficient, pennington2014glove, bojanowski2017enriching} for categorization, wherein plenty of deep learning methods have been adopted, such as TextCNNs~\cite{kim2014convolutional}, RCNNs~\cite{lai2015recurrent}, recurrent neural networks (RNNs)~\cite{liu2016recurrent}, FastText~\cite{joulin2016bag}, BERT~\cite{devlin2018bert}, \emph{etc}. Most of them learn the word representations by firstly projecting the one-hot encoding of each token through a pretrained or randomly initialized word embedding matrices to acquire the dense real-valued vectors, and then feed them into neural models for classification.

These methods, however, have only exploited the low-dimensional semantic representations for each sample text in a supervised way. Some argued that unsupervised latent representations such as topic~\cite{hingmire2014sprinkling,ma2015distributional,li2016news} or cluster modeling~\cite{baker1998distributional, zhang2009text, wang2016semantic, gowda2016semi} mined by latent variable models may be of benefit. \cite{baker1998distributional} maintained that word clustering could deliver the useful semantic information by grouping all words in the corpus and can thus promote the classification accuracy. Moreover, \cite{zeng2018topic} incorporated the neural topic models with Variational Autoencoder (VAE)~\cite{kingma2013auto} into the classification tasks so as to discover the latent topics in the document level and encode the co-occurrence of words with bag-of-words statistics. 

Learning such corpus-level representation can administer to the enrichment of more globally informative features and is thus favorable to the task performance. There are plenty of works adopting VAE for learning these latent variables to boost the text classification performance~\cite{xu2017variational,ayinde2017deep,soares2018effort}.
Nevertheless, there remain problems that we cannot directly treat the sampled latent space of VAE for clustering centroids since there is no mechanism to modulate the representation of different samples towards different mean and variance for a better discrimination purpose under the Gaussian distribution assumption~\cite{lim2020deep}. \cite{song2013auto} and \cite{lim2020deep} alleviate these issues by minimizing the distance between the learnable latent representation from latent variable models and the clustering centers generated from statistical clustering approaches. 

% trained with, in which projecting the word indices into the dense word representations.
Grounding on this, we design an \emph{ad hoc} \textbf{Clu}stering-\textbf{E}nchanced neural model (hereafter CluE) that jointly learns the distributional clustering and the alignment between the domain-aware clustering centroids and word representations in the Euclidean hidden semantic space for text classification, with the vector space assumption that words with similar meanings are close to each other~\cite{mikolov2013distributed}. Instead of directly treating the latent variables as the clustering centroids, we employ a co-adaptation strategy to minimize the difference between the hidden variables and trainable clustering centroids initialized by traditional clustering algorithms with soft alignments.

In the present work, we propose the cluster-token alignment mechanism by assigning relevance probability distribution of clusters to each token, indicating how likely it is that tokens are correlated with each cluster center. In which clustering centroids are co-regulated with learned latent variables and can be regarded as the domain- or task-specific feature indicators.

Our work illustrates that jointly adapting the clustering centroids and learning the cluster-token alignment holds the promise of advancing the text classification performance by incorporating the clustering-aware representations. Our key contributions are:

\begin{itemize} [leftmargin=*]
    \item to empirically and visually demonstrate that the proposed model could surprisingly deliver visually-interpretable text representations for text classification (as fig.~\ref{fig:sent_emb}).
    \item to show that our clustering-token interaction mechanism could apparently capture semantic meanings, including the relevance alignment between clusters and input tokens.
    \item to confirm that our joint learning model eclipses the prevailing baseline models and achieves state-of-the-art results on classifying both short and long texts. 
\end{itemize}{}

\begin{figure}[h] \vskip -5mm
\centering
\begin{minipage}{.45\textwidth}
  \centering
  \includegraphics[width=\linewidth]{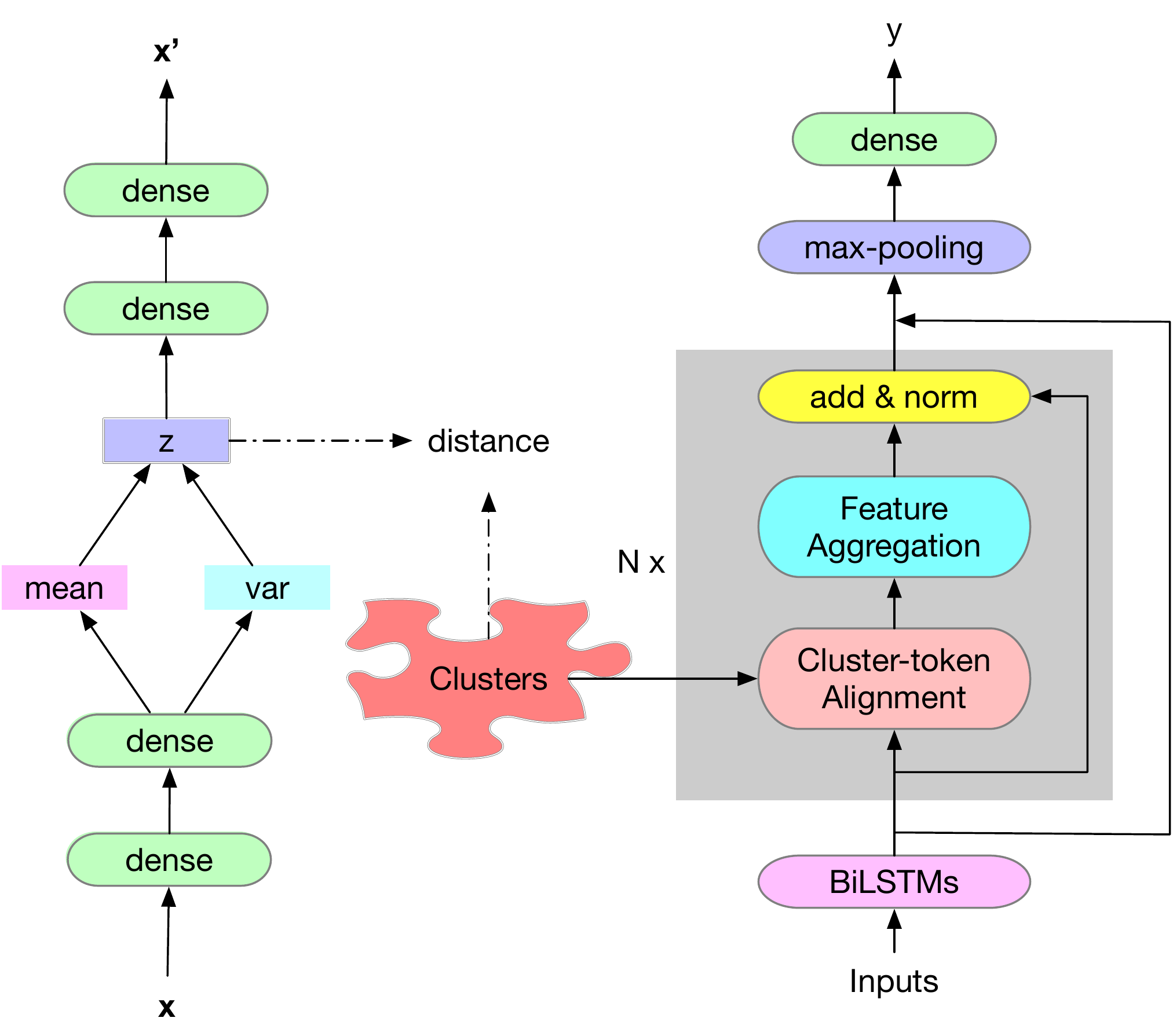} \vskip -2mm
  \captionof{figure}{Illustration of CluE models.}
  \label{fig:CluE}
\end{minipage}%
% \hfill
\hspace{0.2cm}
\begin{minipage}{.41\textwidth}
  \centering
  \includegraphics[width=\linewidth]{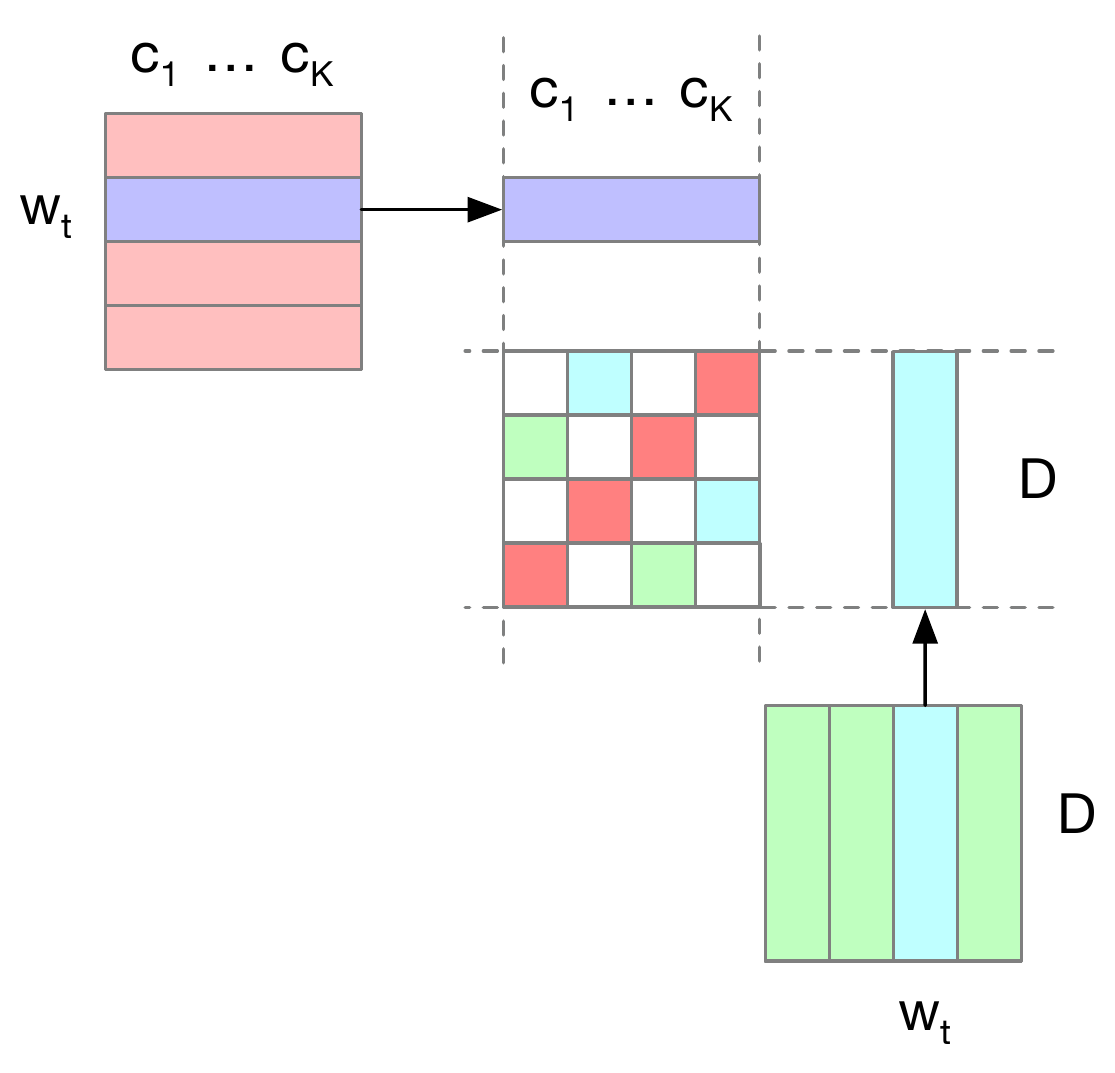} \vskip -2mm
  \captionof{figure}{Aggregation sublayers.}
  \label{fig:agg}
\end{minipage}
\vskip -5mm
\end{figure}

\section{Learning to Cluster and Align}
\label{sec:method}
We propose a clustering-enhanced architecture CluE for text classification (see fig.~\ref{fig:CluE}), consisting of two main components: a cluster co-adaptation module for refining clustering centroids (Sec.~\ref{sec:vae}) and a cluster-enhanced classifier model for categorization (Sec.~\ref{sec:clue}).

We denote the input sequence of words as $\mathbf{w} =\{w_1, w_2, \cdots, w_T \}$ with the length of $T$, the corresponding one-hot representation of labels as $\mathbf{y}$, its \emph{tf-idf} feature as $\mathbf{x}$. To project the discrete word sequences into dense word representations as inputs, we define the word embedding matrices $\mathbf{E} \in \mathbb{R}^{V \times D}$ where $V$ is the vocabulary size of the corpus, and $D$ is the dimension of word embeddings.

\subsection{Distributional Clustering with Hidden Variables}
\label{sec:vae}
Cluster co-adaptation components (left in fig.~\ref{fig:CluE}) leverage latent variable models involving autoencoder (AE) and VAE to conduct the encoding and reconstruction process, simultaneously using the latent representation to regulate the trainable clustering centroids. Typically, we take VAE for the latent variable generation, namely Clustering-VAE, (CVAE), for further explication.

CVAE encodes the \emph{tf-idf} features $\mathbf{x} \in \mathbb{R}^{V}$ for each word sequence into a latent variable word representation $\textbf{z} \in \mathbb{R}^{D} $, then resorts to a modulation mechanism by forcing $\textbf{z}$ to learn together with cluster centers, concurrently reconstructing the original input as  $\mathbf{x}^\prime \in \mathbb{R}^{V}$ from $\textbf{z}$. 
% For readers who are not familar with this, refer to~\cite{kingma2013auto} for detailed explanations.

We denote the trainable cluster centroids consisting of $K$ cluster centers $\{\mathbf{c}_k \in \mathbb{R}^{D} \vert k=1, 2, \cdots, K\}$, where the cluster centroid parameters are initialized with K-means or Gaussian Mixture Models (GMM) trained on the pretrained word representations.

\textbf{Clustering Co-adaptation}\quad 
We apply two-layer feed-forward fully-connected connections with non-linearity, denoted $f$, to project the \emph{tf-idf} features $\mathbf{x}$ into the mean $\pmb{\mu} \in \mathbb{R}^{D}$ and variance $\pmb{\sigma} \in \mathbb{R}^{D}$ under a prior $p(\textbf{z})$ of Gaussian distribution $\mathcal{N}(\pmb{\mu}, \pmb{\sigma})$.
\begin{align}
    \begin{split}\label{eq:enc}
    \mathbf{\mu},  \mathbf{\sigma} = f (\mathbf{x}), \quad
    \mathbf{z} = \pmb{\mu} + \pmb{\sigma} \odot \pmb{\epsilon},
    \end{split}
\end{align}{where $\mathbf{\epsilon} \sim \mathcal{N} (\mathbf{0}, \mathbf{I})$ represents $\pmb{\epsilon} \in \mathbb{R}^{D}$ is sampled from the standard Gaussian distribution $\mathcal{N} (\mathbf{0}, \mathbf{I})$, $\odot$ denotes the element-wise product.}

After sampling $\mathbf{z}$ following the Gaussian prior $p(\textbf{z})$ with the reparameterization trick, we minimize the distance $q_{ik}$ between the $i$-th sample's latent encoding $\mathbf{z}_i$  and each cluster centroid $\mathbf{c}_k$. Here we employ a Student's $t$-distribution kernel following~\cite{xie2016unsupervised} as the distance metric. 
\begin{equation} \label{eq:dec_loss}
    q_{ik} = \frac{(1+\Vert \mathbf{z}_i - \mathbf{c}_k \Vert^2 / \alpha)^{-\frac{\alpha+1}{2}}}{\sum_{k^\prime} (1+\Vert \mathbf{z}_i - \mathbf{c}_{k^\prime} \Vert^2 / \alpha)^{-\frac{\alpha+1}{2}}},
\end{equation}{where $\alpha$ represents the freedom degree of Student's $t$-distribution (we set $\alpha$ as 1 in our experiments). $q_{ik}$ can be considered as the probability assigning the $i$-th sample to the $k$-th cluster.}

The auxiliary target distribution $p_i$ is defined to learn the high confidence assignments~\cite{xie2016unsupervised}. We denote $r_{ik}$ as the squared $q_{ik}$ divided by the total occurrence of each cluster $k$ and $p_{ik}$ as the normalized counterpart among all clusters:

\begin{equation}\label{eq:tgt_dist}
    r_{ik} = \frac{q_{ik}^2} {\sum_i q_{ik}}, \quad p_{ik} = \frac{r_{ik}}{\sum_{k^\prime}r_{ik^\prime}}.
\end{equation}

% \begin{align}
% \begin{split} \label{eq:tgt_dist}
% r_{ik} &{}= \frac{q_{ik}^2} {\sum_i q_{ik}}
% \end{split}\\
% \label{eq:tgt_dist}
% p_{ik} &{}= \frac{r_{ik}}{\sum_{k^\prime}r_{ik^\prime}}
% \end{align}

The objective of clustering co-adaptation is defined as a KL divergence between the the Student's $t$-distribution kernel based distance $q_i$ and the auxiliary target distribution $p_i$:
\begin{equation} \label{eq:loss_cluster}
    \mathcal{L}_\text{cluster}= \sum_i \sum_k p_{ik} \log \frac{p_{ik}}{q_{ik}}.
\end{equation}
\textbf{Feature Reconstruction}\quad 
Meanwhile, the original feature input is reconstructed as  $\mathbf{x^\prime}$ using two feed-forward layers $g$, \emph{i.e.}, $\mathbf{x^\prime} = g(\mathbf{z}) $. The reconstruction component objective $\mathcal{L}_\text{recon}$  is defined as the mean squared error (MSE) between the original features and the reconstructed ones:
\begin{align}
    \mathcal{L}_\text{recon} &{}= \frac{1}{N}\sum_{i=1}^N \Vert \mathbf{x}_i - \mathbf{x^\prime}_i \Vert^2 . \label{eq:recon_loss}
\end{align}
% \begin{align}
% \begin{split}\label{eq:recon}
%     \mathbf{x^\prime} &{}= g(\mathbf{z}) 
% \end{split}\\
%     \mathcal{L}_\text{recon} &{}= \frac{1}{N}\sum_{i=1}^N \Vert \mathbf{x}_i - \mathbf{x^\prime}_i \Vert^2  \label{eq:recon_loss}
% \end{align}{where $N$ denotes the dimension of \emph{tf-idf} features.}

Besides, CVAE employs variational inference to approximate the posterior distribution $q(\mathbf{z} \vert \mathbf{x})$ and force it to approach the prior $p(\mathbf{z})$. The objective of KL diveregence $\mathcal{L}_\text{kld}$ is defined as:
\begin{align} \label{eq:vae_loss}
    \mathcal{L}_\text{kld} = \mathbb{KL} \bigg( q(\mathbf{z} \vert \mathbf{x}) \Vert p(\mathbf{z}) \bigg)  
     \simeq  \frac{1}{2} \sum_{j=1}^D \big( 1 + \log ( (\pmb{\sigma}_j)^2) - (\pmb{\mu}_j)^2 - (\pmb{\sigma}_j)^2 \big),
\end{align}{where $\pmb{\mu}_j$ and $\pmb{\sigma}_j$ represent the $j$-th element of mean and variance vectors respectively. Refer to ~\cite{kingma2013auto} for detailed derivations.}
% \begin{align}
% \begin{split}
%     \mathcal{L}_\text{kld} &{}= \mathbb{KL} \bigg( q(\mathbf{z} \vert \mathbf{x}) \Vert p(\mathbf{z}) \bigg)  
% \end{split}\\ \label{eq:vae_loss}
%      &{}\simeq  \frac{1}{2} \sum_{j=1}^D \big( 1 + \log ( (\pmb{\sigma}_j)^2) - (\pmb{\mu}_j)^2 - (\pmb{\sigma}_j)^2 \big)
% \end{align}{where $\pmb{\mu}_j$ and $\pmb{\sigma}_j$ represent the $j$-th element of mean and variance vectors respectively. Refer to ~\cite{kingma2013auto} for detailed derivations.}

\subsection{Clustering-token Interaction}
\label{sec:clue}
We will further describe the process of clustering-token interaction (the right part in fig.~\ref{fig:CluE}). Given word sequences $\mathbf{w}$ with length $T$, we firstly project the discrete word indices through the embedding matrices $\mathbf{E}$ to acquire the dense representation $\mathbf{H}_e$ as the inputs to the clustering-token interaction module. Afterward, a single-layer bi-directional Long Short-Term Memory (LSTM)~\cite{hochreiter1997long} network to encode the contextual information of tokens into $\mathbf{H}_e$, wherein the output states of forward and backward LSTMs are concatenated at each time step. The dimension of the LSTM hidden state was set to $D/2$, thus we get $\mathbf{H} \in \mathbb{R}^{T \times D}$ after concatenation.
%  \begin{align} 
% \begin{split} \label{eq:embed}
% \mathbf{H}_e &{}= \mathbf{E}(\mathbf{w})
% \end{split}\\
% \begin{split}\label{eq:bi-lstm}
%     \mathbf{H} &{}= \text{bi-LSTM}(\mathbf{H}_e)
% \end{split}
% \end{align}

The clustering-token interaction module is composed of a stack of $N$ identical layers, of which each layer has two sublayers, namely clustering-token global alignment and feature aggregation sublayers. We supplement a residual connection~\cite{he2016deep} around them followed by layer normalization~\cite{ba2016layer} for each stack.

\textbf{Cluster-token Global Alignment}\quad 
The clustering-token alignment sublayer can be described as the alignment score $\mathbf{a} \in \mathbb{R}^{K \times T}$ between the $k$-th cluster $\mathbf{c}_k$ and the $t$-th word representation $\mathbf{h}_t$ in each sequence:
\begin{align}\label{eq:alignment}
    \mathbf{a}_{kt} = \text{align}(\mathbf{c}_k, \mathbf{h}_t)  = \frac{\exp \big( \text{score}(\mathbf{c}_k, \mathbf{h}_t) \big)}{\sum_{k^\prime} \exp \big( \text{score}(\mathbf{c}_{k^\prime}, \mathbf{h}_t) \big)}.
\end{align}
% \begin{align}\label{eq:alignment}
%     \mathbf{a}_{kt} &{}= \text{align}(\mathbf{c}_k, \mathbf{h}_t) \\
%                     &{} = \frac{\exp \big( \text{score}(\mathbf{c}_k, \mathbf{h}_t) \big)}{\sum_{k^\prime} \exp \big( \text{score}(\mathbf{c}_{k^\prime}, \mathbf{h}_t) \big)}
% \end{align}

We adopt the alignment score function as in~\cite{luong2015effective}, which can be interpreted as the content-based relevance function. It measures the relevance between each cluster and each input token:
\begin{align}
    \text{score}(\mathbf{c}_k \mathbf{h}_t)=\left\{
                \begin{array}{ll}
                  \mathbf{c}_k \mathbf{h}_t^\top   & \text{dot}\\
                   \mathbf{c}_k \mathbf{W_a} \mathbf{h}_t^\top  & \text{general}\\
                   \mathbf{W_a} [\mathbf{c}_k; \mathbf{h}_t]  & \text{concat}
                \end{array}
    \right\},
\end{align}{where $\mathbf{W_a} \in \mathbb{R}^{D \times D}$ denotes trainable parameters. Dot alignment score is used in our experiments.}

\textbf{Feature Aggregation} \quad 
The feature aggregation sublayer attends to different dimension of word representations w.r.t each cluster centroids. We refine the $t$-th original representation $\mathbf{h}_t$ using the corresponding cluster-word alignment score $\mathbf{a}_{kt}, k=\{1,\cdots,K\}$ via an Kronecker product. 
\begin{align}
\begin{split}\label{eq:outer}
    \mathbf{u}_{t} &{}= \big( \mathbf{a}_{kt}^\top \otimes \mathbf{h}_t \big) \mathbf{W}_d
\end{split},\\ \label{eq:ln}
    \mathbf{s}_t &{}= \text{layer\_norm} (\mathbf{u}_t + \mathbf{h}_t),
\end{align}{where $\mathbf{u}_{t} \in \mathbb{R}^{T \times D}$ denotes the aggregated output features, $\mathbf{W}_d \in \mathbb{R}^{KD \times D}$ is the trainable weight matrix for dimension reduction,
$\otimes$ is the Kronecker product.}

Fig.~\ref{fig:agg} illustrates the interaction between $\mathbf{a}_{kt}^\top$ (upper-left) and $\mathbf{h}_t$ (lower right corner). So far, we maintain the idea that $\mathbf{a}_{kt}^\top$ calculated from previous sublayer measures the relevance between each cluster (\emph{i.e.},matrix column $\{\mathbf{c}_1,\cdots,\mathbf{c}_K\}$ in the figure) and word tokens(\emph{i.e.}, $w_t$ in the figure).  It can be further interpreted that the resulting matrix in the middle of the figure reveals the relation between cluster and embedding space.   
% token -> cluster
% \begin{figure}[thb]
% \begin{figure}[t]
% \vskip -3mm
% \begin{center}
% \includegraphics[width=\columnwidth]{fig/agg.pdf}
% \vskip -3mm
% \caption{Schematic digram of aggregation sublayer.}
% \label{fig:agg}
% \end{center}
% \vskip -8mm
% \end{figure}

% for each token, token element to each cluster

\label{sec: joint_loss}
\textbf{Classification} \quad 
We define the final representation of stacked clustering-token interaction layers as $\mathbf{S} = \{\mathbf{s}_t \vert t=\{1,2,\cdots, T\} \}$. Firstly we add a residual connection from after the bi-LSTMs followed by a layer normalization, and pass it through a dense layer with a non-linearity. Then a max-pooling layer is employed to get the final text representation $\mathbf{o} \in \mathbb{R}^{D}$, with a linear projection to output the final classification logits $\mathbf{\hat{y}}$.
\begin{align}
\begin{split}\label{eq:pooling}
    \mathbf{o} &{}= \max_{t=1}^T \text{ReLU} (\text{layer\_norm} (\mathbf{S} + \mathbf{H}) \mathbf{W}_O)
\end{split},\\
\begin{split}\label{eq:pred}
\mathbf{\hat{y}} &{}= \text{softmax}(\mathbf{o} \mathbf{W}_p)
\end{split},\\ \label{eq:cls_loss}
\mathcal{L}_\text{cls} &{}= - \sum_i \mathbf{y}_i \log \mathbf{\hat{y}}_i,
\end{align}{where $\mathbf{W}_O \in \mathbb{R}^{D \times D^\prime}$ and $\mathbf{W}_p \in \mathbb{R}^{D^\prime \times \vert C \vert}$ are trainable parameters, $D^\prime$ denotes the hidden dimension, $|C|$ is the number of classes. 
$\mathcal{L}_\text{cls}$ represents the cross entropy loss between label $\mathbf{y}$ of one-hot encoding and the output logits $\mathbf{\hat{y}}$ for the classifier.}

\textbf{Joint Training} \quad 
Finally, the joint training objective of the CluE model is designed as:
\begin{align} 
\begin{split}\label{eq:loss}
    \mathcal{L} = \mathcal{L}_\text{cls} + \lambda_1 \mathcal{L}_\text{cluster} + \lambda_2 \mathcal{L}_\text{recon}+ \lambda_3 \mathcal{L}_\text{kld}
\end{split},
\end{align}{where $\{\lambda_1,\lambda_2,\lambda_3\}$ represent scalars for scaling and are selected on holdout sets.}

\section{Experiment Settings}
Our experiments are conducted on the task of text classification using four different benchmark datasets for short and long text classification separately. For comparison, we report the performance of a bunch of strong baseline models w.r.t the prediction accuracy on the test set. Code in TensorFlow~\cite{abadi2016tensorflow} will be available after the double-blind review (for reviewers, please check the attached materials).

\label{sec:exp}
\subsection{Data}

% \usepackage{booktabs}
% \usepackage{multirow}
% \usepackage{graphicx}
\begin{table}
\centering
\resizebox{.8\textwidth}{!}{%
\begin{tabular}{@{}lcccccc@{}}
\toprule
\textbf{Type} &
  \textbf{Datasets} &
 \textbf{Classes} &
  \textbf{Train} &
  \textbf{Test} &
  \textbf{Vocabulary} &
  \textbf{Avg. $T$} \\ \midrule
\multirow{4}{*}{short} & AG's News & 4  & 120k    & 7.6k    & 13,464  & 7   \\
                       & DBpedia   & 14 & 560k    & 70k     & 35,165  & 3   \\
                       & Amazon    & 5  & 3,000k  & 650k    & 47,638  & 5   \\
                       & Yahoo     & 10 & 908,904 & 227,227 & 35,308  & 11  \\ \midrule
\multirow{4}{*}{long}  & AG's News & 4  & 120k    & 7.6k    & 25,537  & 32  \\
                       & DBpedia   & 14 & 560k    & 70k     & 132,251 & 48  \\
                       & IMDB      & 2  & 40k     & 10k     & 38,272  & 241 \\
                       & R8        & 8  & 5,485   & 2,189   & 5,160   & 66  \\ \bottomrule
\end{tabular}%
}\vskip 2mm
\caption{Summary of datasets after preprocessing.}
\label{tab:data}
\vskip -4mm
\end{table}

Table~\ref{tab:data} summarizes the statistics of different datasets. Original train/test sets have remained except 80/20 train/test split on IMDB and Yahoo datasets. 10\% of the training data is extracted as holdout sets. The preprocessing is composed of handling web links, digits, punctuations, and lowercasing after replacing with \textsc{[UNK]} placeholders tokens whose occurrences are not greater than 3. 
% On Amazon Reviews and Yahoo! Answers, we only classify their short text fields due to the infeasibility of training baseline models such as TMN and TextGCN (described in sec.~\ref{sec:exp_models}) on such large volumes of data.

We adopt the title texts from AG's News\footnote[4]{\label{note_data} These datasets are obtained from \cite{zhang2015character}}, DBpedia\footnoteref{note_data}, Amazon Reviews\footnoteref{note_data} and Yahoo! Answers\footnote[3]{Available online at \url{https://webscope.sandbox.yahoo.com/catalog.php?datatype=l&did=11}} text data for short text classification. 
Meanwhile, content fields from AG's News\footnoteref{note_data} and DBpedia\footnoteref{note_data}, as well as IMDB movie reviews\footnote[5]{ \url{https://www.kaggle.com/lakshmi25npathi/imdb-dataset-of-50k-movie-reviews}. } and R8\footnote[6]{Available online at~\url{https://www.cs.umb.edu/~smimarog/textmining/datasets/}} datasets are utilized as long texts\footnote[7]{Long texts from Amazon and Yahoo datasets are not adopted due to the infeasibility to train models like TMN and TextGCN on such large volumes of data.}.

% \paragraph{AG's News\footnote[4]{\label{note_data} These datasets are obtained from \cite{zhang2015character}}} It contains 120,000 and 7,600 samples for training and testing out of 4 categories, in which each sample includes both the news title and content texts for short and long text classification.
% \paragraph{DBpedia^{\footnotemark[\ref{note_data}]}} DBpedia dataset is constructed by extracting title (\emph{i.e.}, short) and abstract (\emph{i.e.}, long) from Wikipedia articles. It has 560,000 training samples and 70,000 testing samples from 14 ontology categories.

% \paragraph{Amazon Reviews^{\footnotemark[\ref{note_data}]}}  The Amazon reviews dataset consists of 3.0 million training samples and 0.65 million test samples from one-to-five rating labels. The fields include review titles and review contents.

% \paragraph{Yahoo! Answers\footnote[3]{Available online at \url{https://webscope.sandbox.yahoo.com/catalog.php?datatype=l&did=11}}} The raw dataset has 4,483,032 question titles and their corresponding contexts. We collect from the top 10 largest main categories and randomly split them into 80/20 as the training and test set. 

\subsection{Models}
\label{sec:exp_models}
% We conduct the comparison experiments on following mainstream baseline models.

\textbf{Comparison Models} \quad We test mainstream baseline models including TextCNN\footnote[2]{\label{fn1}indicates we obtain the source code from the author and use its default settings for experiments.}, BiLSTM, RCNN~\cite{lai2015recurrent} and AttBiLSTM~\cite{lin2017structured}. 6 and 12 Transformer~\cite{vaswani2017attention} encoder layers followed by a max-pooling layer are implemented. Topic Memory Network (TMN)\footnoteref{fn1}~\cite{zeng2018topic} that combines neural topic model and memory network to tackle topic modeling and classification jointly were experimented. To further compare with the recently prevailing Text Graph Convolutional Networks \footnoteref{fn1} (TextGCN)~\cite{yao2019graph}, which builds corpus-level graph using word co-occurrences and document word relations, we conduct experiments on the source code. Additionally, we reproduce a text-level GCN based on the dependency parsing graph for each input sample, similar to~\cite{zhang2019aspect}.

% \subsubsection{Comparison Models}
% \textbf{TextCNN}\footnote[2]{\label{fn1}indicates we obtain the source code from the author and use its default settings for experiments.}~\cite{kim2014convolutional}  \quad It integrates Convolutional Neural Networks(CNN) on the sequence direction followed by global max-pooling for text classification.

% \textbf{BiLSTM} \quad It employs single-layer Bi-LSTM to capture sequential features as contextual representations.

% \textbf{RCNN}~\cite{lai2015recurrent}\quad  Bi-LSTM encoded states and word embeddings are concatenated to capture word context embeddings.

% \textbf{AttBiLSTM}~\cite{lin2017structured}\quad  Attention mechanisms are supplemented after Bi-LSTMs to better capture sentence-level representations.

% \textbf{Transformer}~\cite{vaswani2017attention}\quad  6 and 12 layer Transformer encoder stacks followed by a max-pooling layer are implemented.

% \textbf{TMN}\footnoteref{fn1}~\cite{zeng2018topic}\quad  Topic Memory Network (TMN) combines neural topic model and memory network to tackle topic modeling and classification jointly. 

% \textbf{TextGCN}\footnoteref{fn1}~\cite{yao2019graph}\quad  Text Graph Convolutional Networks (TextGCN) builds corpus-level graph using word co-occurrences and document word relations. Besides, we reproduced a text-level GCN based on the dependency parsing for each input sample, similar to~\cite{zhang2019aspect}.

\textbf{Proposed Models} \quad  We run our model with three experimental settings: \textbf{CluE-baseline}, which directly use clusters initialized with K-means method for CluE training by removing the latent variable components like CVAE; \textbf{CluE-CAE}, which applies Autoencoder (AE) Networks based on CluE-baseline to generate hidden states, namely Clustering-AE (CAE);  \textbf{CluE-CVAE} supplemented on CluE-baseline as described in sec.~\ref{sec:method}.
% \subsubsection{Proposed Models}

% \textbf{CluE-baseline} \quad For comparison, we directly use clusters initialized with the K-means method for CluE training by removing the latent variable components like CVAE.

% \textbf{CluE-CAE} \quad We apply Autoencoder (AE) Networks based on CluE-baseline to generate hidden states, namely Clustering-AE (CAE).

%  \textbf{CluE-CVAE} \quad The proposed CVAE are supplemented on CluE-baseline to construct CluE-CVAE. 
%  (as described in sec.~\ref{sec:method}).

\textbf{Hyperparameter settings} \quad  Hyper-parameters are tuned with grid search on both short and long texts. We use the 300-dimensional pretrained Glove embeddings\footnote[3]{ Available at \url {http://nlp.stanford.edu/data/glove.6B.zip} } to initialize the word embedding matrix and clustering centroids. The maximum sequence length is set to 20 for all short texts except Yahoo dataset with 28 and the counterpart for long texts are 120 for AG's News and DBpedia and 200 for Yahoo! Answers and Amazon Reviews dataset.
For the optimization, we use Adam optimizer~\cite{kingma2014adam} with the learning rate 1e-3, batch size of 512 and 64 for long and short inputs, respectively. The training process is set as a maximum of 10,000 steps with early stopping patience of 30 steps. Gradients are clipped when the L2 norm is more than 5. To avoid model over-fitting, we set the dropout rate as 0.2 during training.
% \footnote[8]{Refer to our public code for further details}

% \usepackage{booktabs}
% \usepackage{multirow}
% \usepackage{graphicx}
\begin{table}
\centering
\resizebox{\textwidth}{!}{%
\begin{tabular}{@{}llllll|llll@{}}
\toprule
\multicolumn{2}{l}{\multirow{2}{*}{}} & \multicolumn{4}{c|}{short text} & \multicolumn{4}{c}{long text} \\ \cmidrule(l){3-10} 
\multicolumn{2}{l}{} & AG's News & DBpedia & Yahoo & Amazon & AG's News & DBpedia & R8 & IMDB \\ \midrule
TextCNN~\cite{kim2014convolutional} &  & 0.8696 & 0.7049 & 0.6369 & 0.4762 & 0.9122 & 0.9864 & 0.9571* & 0.9023 \\
BiLSTM &  & 0.8622 & 0.6983 & 0.6364 & 0.4683 & 0.9064 & 0.9839 & 0.9631* & 0.8976 \\
RCNN~\cite{lai2015recurrent} &  & 0.8636 & 0.7020 & 0.6397 & 0.4730 & 0.9109 & 0.9867 & 0.9719 & 0.9067 \\
AttBiLSTM~\cite{lin2017structured} &  & 0.8682 & 0.6997 & 0.6366 & 0.4775 & 0.9082 & 0.9858 & 0.9604 & 0.8954 \\
\multirow{2}{*}{Transformer~\cite{vaswani2017attention}} & \multicolumn{1}{l|}{L6} & 0.8649 & 0.6816 & 0.6229 & 0.4630 & 0.8904 & 0.9794 & 0.9357 & 0.8151 \\
 & \multicolumn{1}{l|}{L12} & 0.8661 & 0.6710 & 0.6178 & 0.4654 & 0.8913 & 0.9693 & 0.9301 & 0.8242 \\
TMN~\cite{zeng2018topic} &  & 0.8537 & 0.5632 & 0.5854 & - & - & - & - & - \\
TextGCN~\cite{yao2019graph} &  & 0.8572 & - & - & - & - & - & 0.9707* & - \\
Text-level GCN &  & 0.8652 & 0.6980 & 0.6357 & 0.4689 & 0.9051 & 0.9847 & 0.9554 & 0.8761 \\ \midrule
CluE-baseline &  & 0.8821 & 0.7053 & 0.6410 & 0.4318 & 0.9150 & 0.9867 & 0.966 & 0.8709 \\
CluE-CAE &  & 0.8832 & 0.7039 & 0.6446 & 0.4782 & 0.9146 & 0.9873 & 0.9641 & 0.8996 \\
CluE-CVAE &  & \textbf{0.8846} & \textbf{0.7055} & \textbf{0.6475} & \textbf{0.4884} & \textbf{0.9198} & \textbf{0.9887} & \textbf{0.9738} & \textbf{0.9096} \\ \bottomrule
\end{tabular}%
} \vskip 2mm
\caption{Summary of model performance (accuracy on the test set) on classification benchmark datasets, where `*' indicates the result is from~\cite{yao2019graph}.}
\label{tab:all_acc} \vskip -7mm
\end{table}

\section{Results}
\label{sec:res}
% Further explanations and analysis will be given in this section.

\subsection{Quantitative Results}
% we found that adding sinusoidal transformer-like positional embedding hurts the performance of our models on the test set.

Table~\ref{tab:all_acc} exhibits the classification performances of described models (sec.~\ref{sec:exp_models}) measured in test accuracy. It is obvious from the table that proposed CluE models outperform various dominant baseline models by a clear margin in different benchmark datasets involving both short and long texts.

As shown in the table, CNN-dominant models, \emph{i.e.}, TextCNNs, perform steadily well on all datasets while LSTM-based models including Bi-LSTM, RCNN, and AttBiLSTM achieve better results on long texts. This is consistent with our initial intuition.
% Our implemented 6-layer and 12-layer Transformers without pretraining 

The performance of prevailing GCN models, \emph{i.e.}, TextGCN and Text-level GCN, matches that of LSTM- or CNN- dominant methods. However, TextGCN suffers from the lack of transferability and memory-efficiency, making it difficult to apply corpus-level TextGCNs on large datasets. 

It is empirically shown that combing topic networks and memory networks to construct TMN models dramatically slows down the training process and can not be afforded for large short text datasets, even not to mention long texts. Notably, `-' symbols in table~\ref{tab:all_acc} represent the inacquirability of experimental results. Specifically, the overlong time costs (\emph{e.g.}, $\approx$4600 seconds per training epoch out of maximum 800 epochs on single NVIDIA Titan RTX GPU for Amazon short datasets) for TMN and memory/running errors in the corpus-level graph construction for TextGCN when handling large volumes of data.

\textbf{Impact of CluE Architectures}\quad 
Table~\ref{tab:all_acc} illustrates the comparison between our CluE models with different settings for classifying AG's News short texts. Our CluE-baseline, CluE-CAE and CluE-CVAE models achieve 2.31\%, 2.44\% and 2.6\% improvement on the test accuracy compared with CluE-baseline without clustering-token interaction layers, \emph{i.e.}, bi-LSTM (see fig.~\ref{fig:CluE} for schematic intuition).

\begin{figure}[thb] 
\vskip -4mm
\centering
  \begin{subfigure}[b]{0.23\linewidth}
    \includegraphics[width=\textwidth]{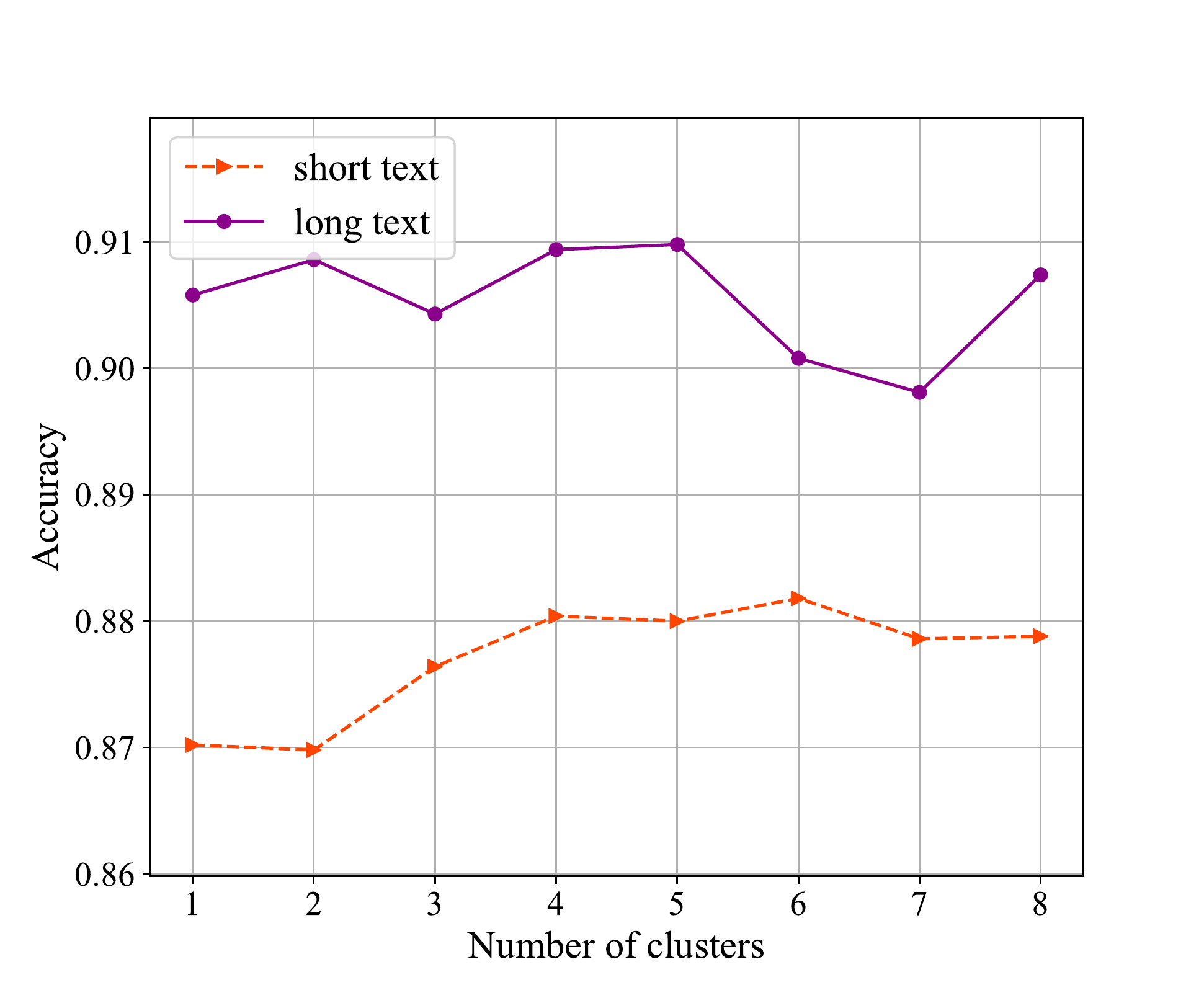} \vskip -2mm
    \caption{Number of clusters.}
    \label{fig:n_clusters}
  \end{subfigure}
  \begin{subfigure}[b]{0.23\linewidth}
    \includegraphics[width=\textwidth]{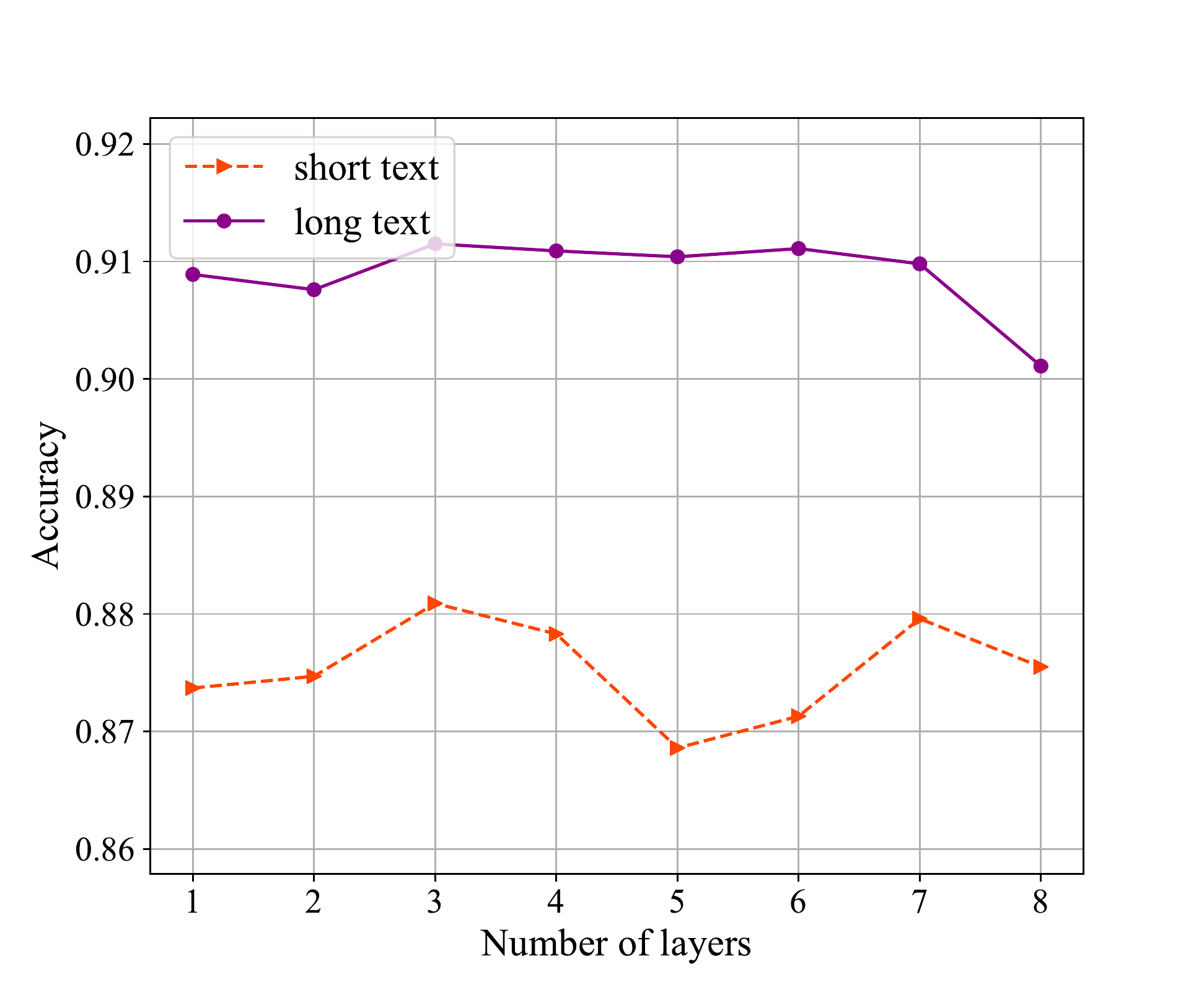}  \vskip -2mm
    \caption{Number of layers.}
    \label{fig:n_layers}
  \end{subfigure} 
    \begin{subfigure}[b]{0.23\linewidth} 
    \includegraphics[width=\textwidth]{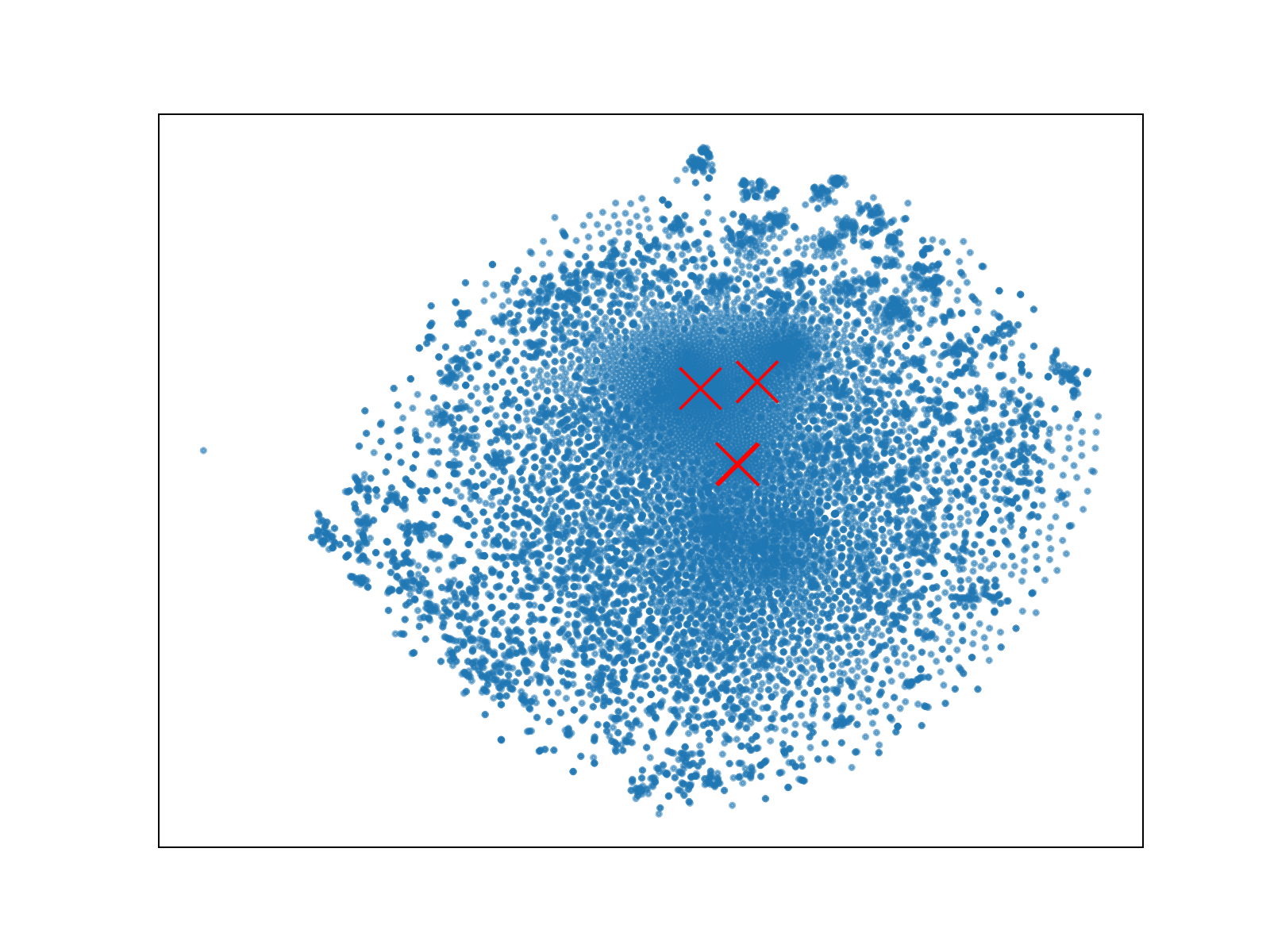}
    \caption{}
    \label{fig:c_embed}
  \end{subfigure}
  \begin{subfigure}[b]{0.23\linewidth}
    \includegraphics[width=\textwidth]{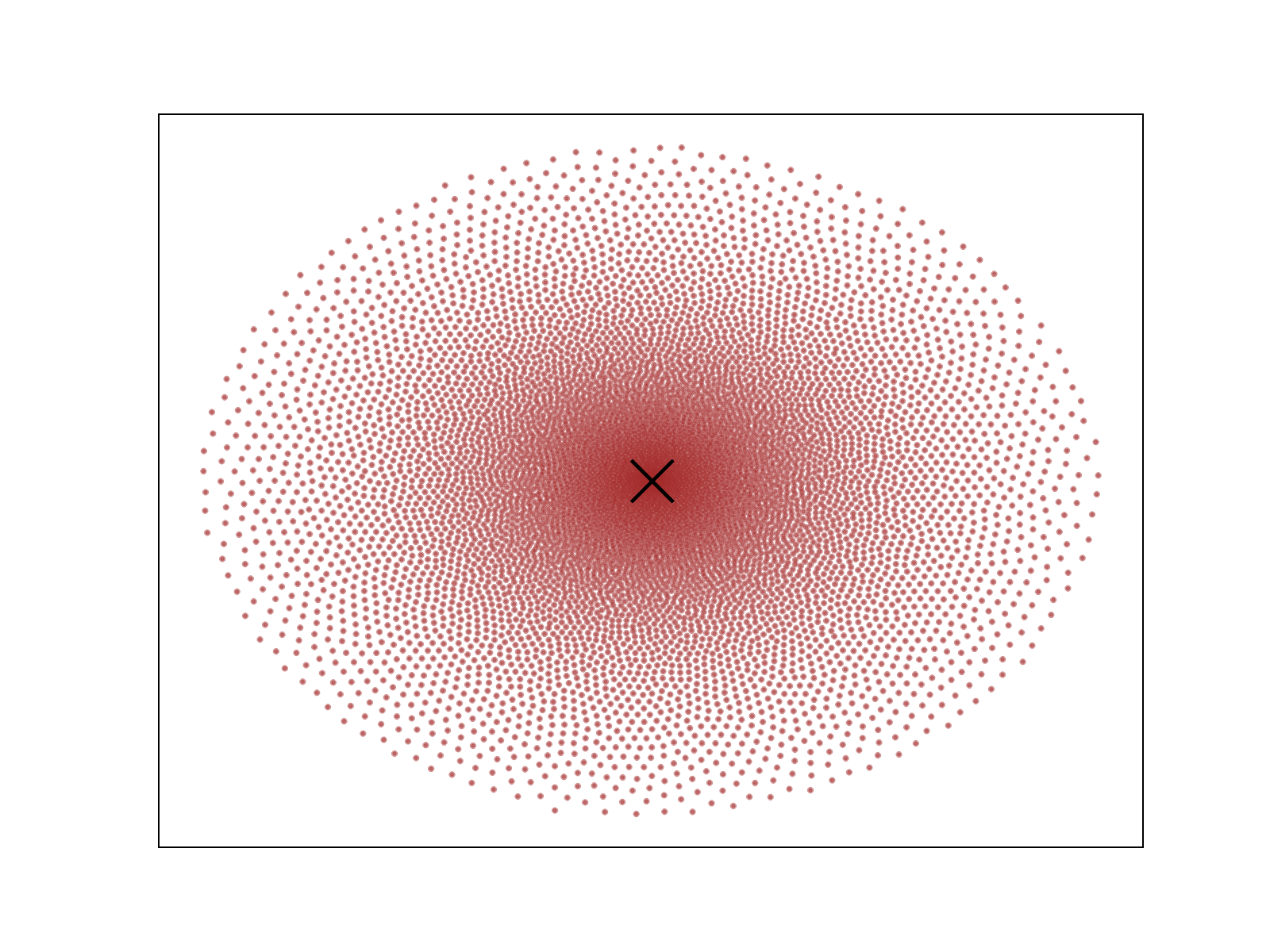}
    \caption{}
    \label{fig:cluster_z}
  \end{subfigure} %\vskip -2mm
  \caption{(a) and (b) reports the accuracy on test set with the increase of the number of clusters (a) or that of layers (b); (c) and (d) are scatter visualization of learned cluster centroids (`x' marks) and trained word embeddings (drawn as blue points in (c)) or latent hidden variables $\mathbf{z}$ (drawn as brown points in (d)) using t-SNE on AG's News dataset. }
    \label{fig:plot}
    \vskip -5mm
\end{figure}

Further, we evaluate the impact of the number of layers and clusters ranging from 1 to 8 on CluE-CVAE models and plot a line graph in fig~\ref{fig:plot} using AG's News dataset.

\textbf{Impact of Cluster Numbers} \quad  It can be seen from fig.~\ref{fig:n_clusters} that the model performance remains stable when the range of layer number is within 1 to 5, and then decreases afterward when classifying long texts. In contrast, the curve of short texts initially shows an increasing trend and reach a plateau from 4 to 8. It is observed that the model performance is superior when the number of clusters is approaching to that of classes. In our optimal settings of CluE-CAE and CluE-CVAE, the optimal cluster counts are both 4, which is exactly the class number of AG's News. Such curves could match our intuition that clustering mechanisms might learn the semantic meanings beneficial for classification to some extent.  
% With the law of Occam's razor, we assume that setting the cluster number as that of classes could save a lot of computing power.

\textbf{Impact of Layer Numbers} \quad 
Fig.~\ref{fig:n_layers} witnesses the influence trend of layer numbers. We find that models with three layers reach their acme on test performance for short texts whilst the test accuracy with layer numbers range from 3 to 7 remains a steady stage for long texts. It shows that models with layer number 3 can get superior results. We recommend readers not to increase the depth of layers too much according to the law of Occam's razor.

% \begin{figure}[thb] \vskip -2mm
% \centering
%   \begin{subfigure}[b]{0.42\linewidth}
%     \includegraphics[width=\textwidth]{fig/n_clusters.pdf}
%     \caption{Number of clusters.}
%     \label{fig:n_clusters}
%   \end{subfigure}
%   %
%   \begin{subfigure}[b]{0.42\linewidth}
%     \includegraphics[width=\textwidth]{fig/n_layers.pdf}
%     \caption{Number of layers.}
%     \label{fig:n_layers}
%   \end{subfigure} \vskip -2mm
%   \caption{The line graph reports the accuracy on test set with the increase of the number of clusters (left) or that of layers (right).}
%     \label{fig:impact_line}
%     \vskip -5mm
% \end{figure}

\subsection{Qualitative Analysis}
To illustrate the impact of CluE-CVAE models, we will display visualizations on AG's News test set unless mentioned otherwise. Main model settings used for visualization are as follows: clustering number 4, clustering-token interaction layer number 4, scaling factors $\lambda_1 = \lambda_2 = \lambda_3 = 1$.

\begin{figure}
\vskip -6mm
% [thb]\vskip -5mm
\centering
\begin{minipage}{.45\textwidth}
  \centering
  \includegraphics[width=\linewidth]{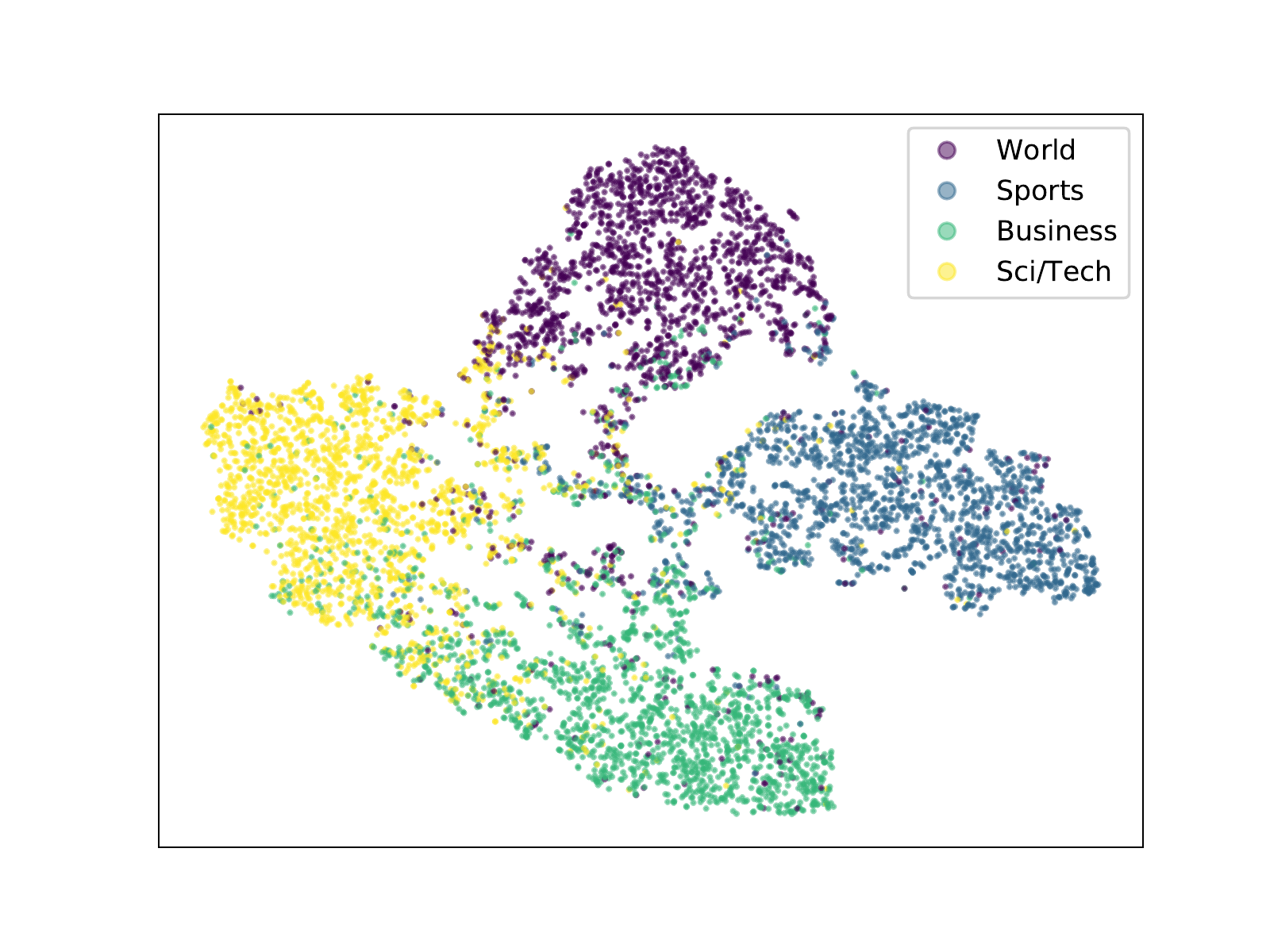} \vskip -2mm
  \captionof{figure}{t-SNE visualization for the learned text representations of proposed CluE models, wherein text representations with different labels are drawn in different colors.}
  \label{fig:sent_emb}
\end{minipage}%
% \hfill
\hspace{0.4cm}
\begin{minipage}{.4\textwidth}
  \centering
  \includegraphics[width=\linewidth]{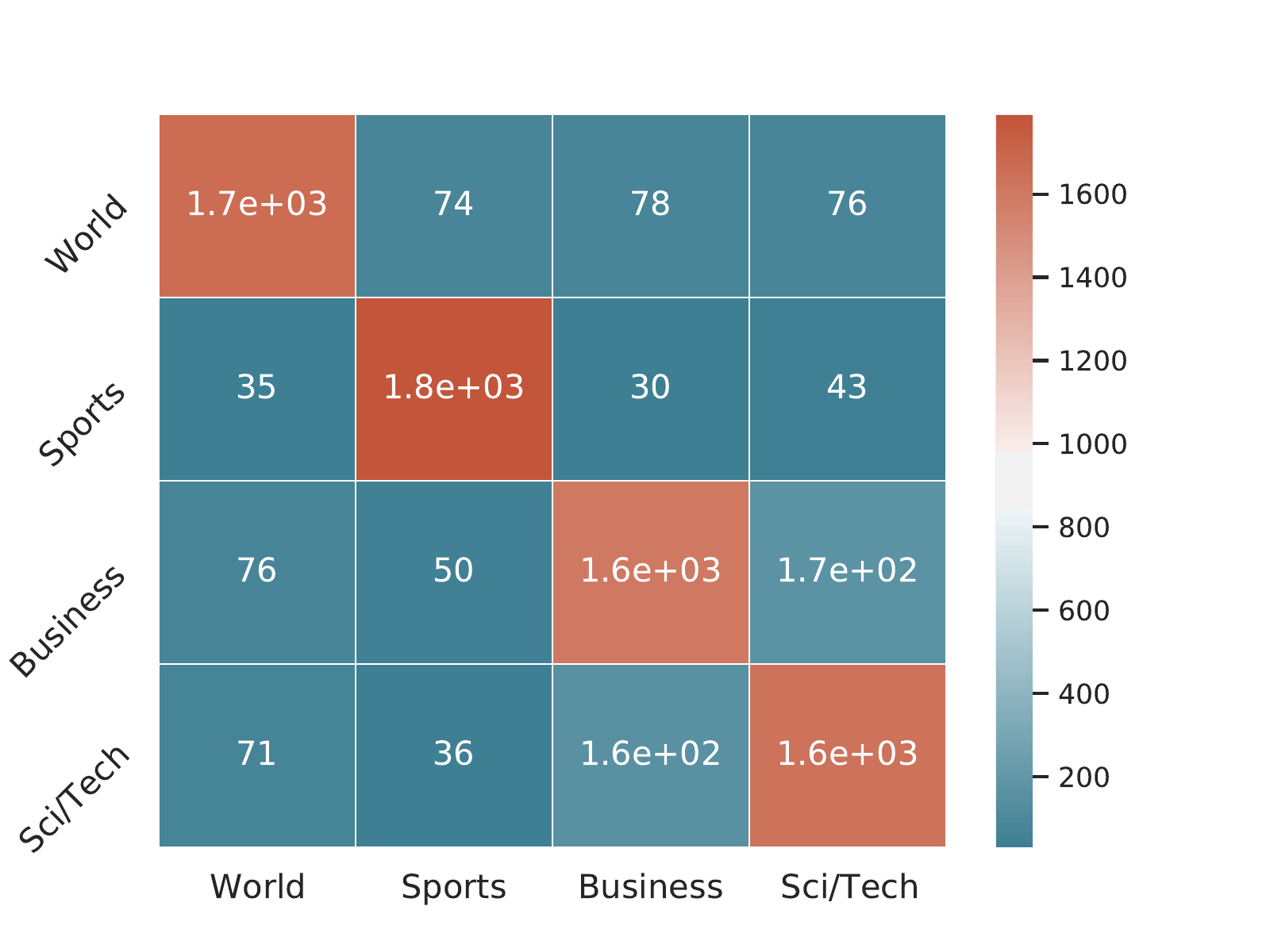} \vskip -2mm
  \captionof{figure}{Plot of confusion matrix to evaluate the classification accuracy, where rows and columns denote the actual and predicted labels, the annotation values are the number of samples.}
  \label{fig:confusion_mat}
\end{minipage}
\vskip -6mm
\end{figure}

\textbf{Text Representation Visualization} \quad 
In the proposed CluE-CVAE model, sentence embeddings are acquired after the max-pooling layer (upper-right in fig.~\ref{fig:CluE}). We plot the sentence embedding together with its corresponding classes into a scatter graph in fig.~\ref{fig:sent_emb}. It is dramatically vivid that sentence representations from different label classes are apparently partitioned into four distinct clusters and well separated from each other, hugely validating the ideas that our CluE models deliver the better discrimination between different classes. Meanwhile, such orthogonal clustering groups could provide more informative meanings to classifiers and thus boost the classification performance. Notably, we can observe from the plot that the data points near the intersection between the region of ``Business'' (in yellow) and ``Sci/Tech'' (in green) have plenty of mixed points. We make an extrapolation that this is due to the disambiguity of samples from these two classes. We will testify this at the end of this section.

\textbf{Clustering Centroid Visualization} \quad  We plot the scatter graph of learned clustering centroids with trained word embeddings (fig.~\ref{fig:c_embed}), and those with all hidden variables of each input texts (fig.~\ref{fig:cluster_z}). We find that word embeddings after training circumfuse cluster centroids and have a high density around clustering centers. This underpins the proposed clustering-token interaction mechanism, showing that word representations move towards clustering centers to be more classifier-discriminative and class-informative in the compact cluster-token high-dimensional vector space, though no direct minimization mechanism in between were conducted.
% clustering fig

Fig~\ref{fig:cluster_z} witnesses that cluster centers are located at near the center of the multi-variance Gaussian distribution that latent variables $\mathbf{z}$ obey, indicating that our KL divergence loss (eq.~\ref{eq:loss_cluster}) for minimizing the distance between hidden variables and cluster centers works well.

\begin{figure}[thb] 
\vskip -3mm
\centering
\begin{subfigure}[b]{\textwidth}
   \includegraphics[width=1\linewidth]{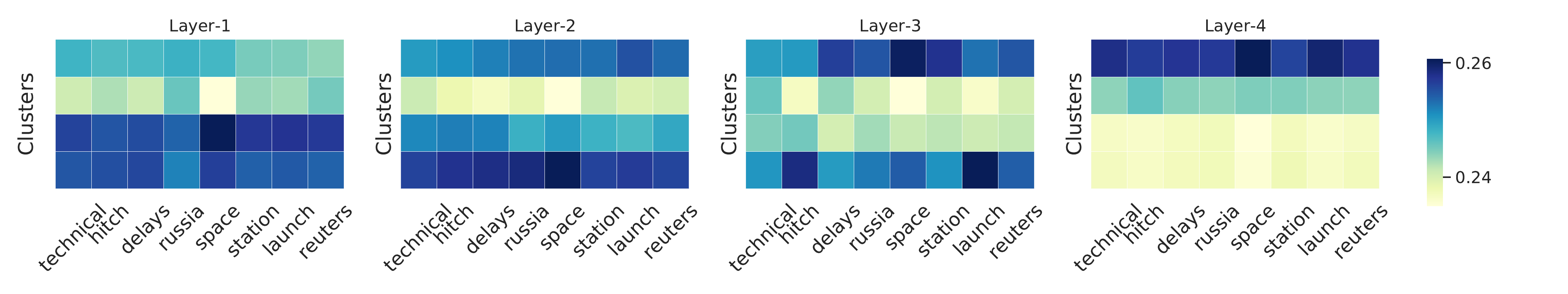}
%   \caption{}
%   \label{fig:align_0} 
\end{subfigure}
\vskip -2mm
\begin{subfigure}[b]{.6\textwidth}
   \includegraphics[width=1\linewidth]{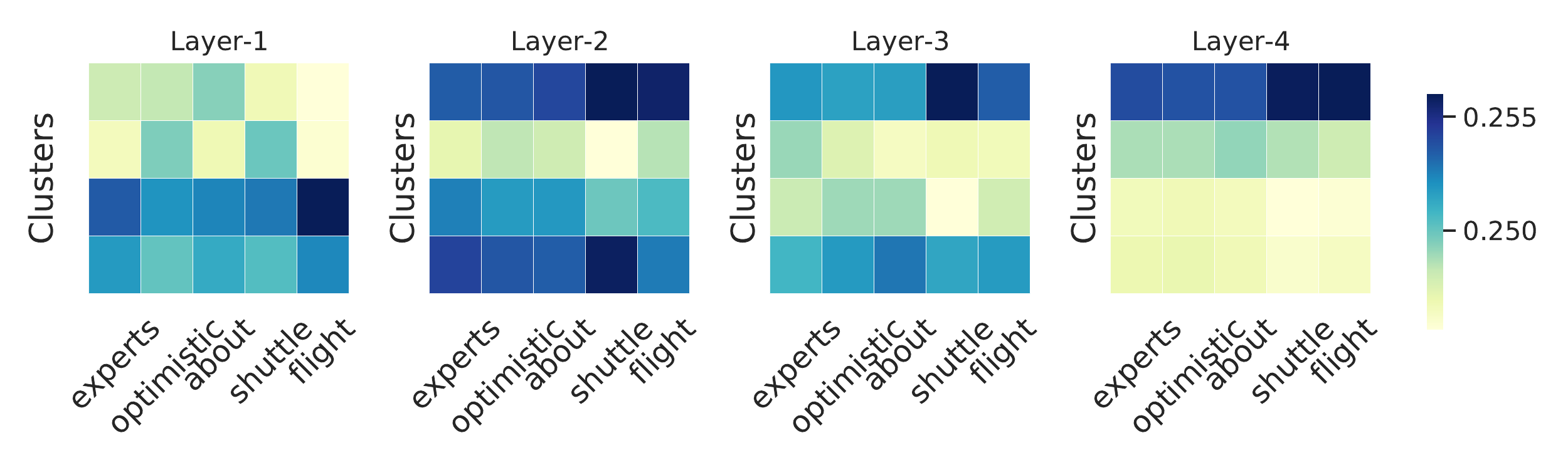}
%   \caption{}
%   \label{fig:align_1}
\end{subfigure}
\vskip -2mm
\caption[]{Visualization of outputs from each clustering-token alignment sublayer for example sentences ``technical hitch delays russia space station launch reuters'' and ``experts optimistic about shuttle flight''. 
The x-axis and y-axis of each plot correspond to tokens and clusters respectively. Each pixel shows the relevance weight $\alpha_{kt}$ between the $k$-th cluster (row) and $t$-th token (column) (see eq.~\ref{eq:alignment}).}
\label{fig:align}
\vskip -3mm
\end{figure}

\textbf{Cluster-token Alignment Visualization} \quad 
It is evident from the heatmap visualization (fig.~\ref{fig:align}) that the relevance for clusters progressively increases or decreases with the depth increase of layers, finally focusing on a single cluster to form the final sentence representations (fig.~\ref{fig:sent_emb}). 
% We put the alignment plots of final layers on more example texts in Appendices~\ref{sec: ap-alignmap}.
% alignment fig c x D visualization

% 2/3 class
%  cluster final layers concentrate on single columns

\textbf{Error Analysis} \quad 
\label{par:error}
% pick some cases that can learned in our model but do not in baselines
% weak between 2/3 classes -> fig 1
Fig.~\ref{fig:confusion_mat} displays the confusion matrix between the golden and predicted labels, in which samples of label ``Business'' (hereafter `B') and ``Sci\slash Tech''(hereafter `S') are mis-predicted in the most cases. 

To unveil this, we go through the wrongly predicted samples and attribute this to three reasons: true mispredictions, mislabels, or ambiguities between these two classes.

Firstly, we observe that samples involving tech company or product names could be predicted as `S' by mistake, such as ``google puts desktop search privacy up front'', ``amazon sues spammers for misleading consumers'', ``ibm quitting computer business''.

Some are labeled by mistake but predicted correctly, such as:
\setlist{nolistsep}
\setlist[itemize]{leftmargin=*}
\begin{itemize} [noitemsep]
  \item labeled `B' but predicted as `S': ``fcc mulls airborne mobile phone use'', ``web based kidney match raises ethics questions''.
  \item labeled `S' but predicted as `B': ``google founders selling off stock'', ``ibm buys two danish services firms'', ``'microsoft signs two indian deals'''.
\end{itemize}

However, there also exist plenty of ambiguous samples, like ``google targets software giant'', ``google puts desktop search privacy up front'', ``red hat replaces cfo'' which are labeled as `B' in the dataset.

This exactly matches our previous assumption, according to fig.~\ref{fig:sent_emb}. It can be further inferred that text representations learned by our models could surprisingly be consistent with our intuition in terms of the semantic vector space, validating the impact of the proposed models. 

% \section{Related Work}
% \label{sec:bg}
% \subsection{Text Classification}

% \subsection{Clustering Methods}

\section{Conclusion}
The proposed CluE architecture enjoys both the advantage of latent variable models and clustering-token interaction mechanisms without introducing additional knowledge from outside, which allows for attending to semantic meanings from both tokens and their corresponding embedding elements. It overshadows the performance of prevailing baseline models and empirically proves that the proposed clustering-token interaction mechanism could be of benefit for achieving informative sentence embeddings. In the future, we will expand our experiments on the other NLP tasks such as natural language inference and neural machine translation.

\section*{Broader Impact}
%  ethical aspects 
%  future societal consequences

% discuss both positive and negative outcomes, 
% a) who may benefit from this research
% b) who may be put at disadvantage from this research 
% c) what are the consequences of failure of the system
% d) whether the task/method leverages biases in the data.

This work has the following potential positive impact on society:
\begin{itemize}
    \item Leveraging the proposed clustering enhanced method, user-generated text data can be analyzed and categorized to interpret social behaviors of users, such as rumor detection, spam detection, hate-speech detection, and sentiment analysis. In addition, these cluster-based analyses can also be embedded into recommender systems for the purpose of user-profiles and recommendations.
    \item Latent variable models and clustering mechanisms have been leveraged to boost the prediction accuracy on text classifiers, including intent detection, Twitter hashtag prediction, sentiment analysis, \emph{etc}. Latent variable models are shown to be effective in enriching representations in an unsupervised way.
    \item Proposed cluster-token interaction mechanism could learn more informative and meaningful text representations as visualized in the paper, which can be adopted in other relevant applications such as natural language inference, topic modeling, \emph{etc}.
    \item Our methods connect between the neural clustering approaches and text classification tasks, which may be widely applied in the industry. Besides, other prevailing methods could also borrow this idea, like enriching BERT families with similar methods.
    \item This clustering-enhanced method can be transferred into semi-supervised learning in practice by providing pseudo labels by assigning learned text representations to the closest cluster of which label is defined as the mode. This may reduce the requirement of amounts of golden labeled data and thereby curtail the cost.
\end{itemize}

At the same time, this work may have some negative consequences because there is no mechanism to enforce clustering centroids to be differential and far apart with each other, which can be seen from our plots. The performance may be further promoted by handling this problem.

It should be noted that the failure of the system could result in the wrong prediction of classifications, which may further lead to accumulative errors in downstream tasks, such as spoken dialogue systems.  

The current study may suffer from implicit biases in NLP systems:
\begin{itemize}
    \item the demographic/representative bias that might be learned implicitly from a certain domain.  Take our experimental AG's News dataset, for example, texts containing technical company names may be possibly predicted into ``Sci/Tech'' categories. 
    \item the implicit bias from word embeddings pretrained on Wikipedia. For example, family- or gender-related topics occurred more about women. Although Wikipedia is a major dataset used for pretraining embeddings, it has inevitable biases with crowdsourcing.
\end{itemize}

As for the ethical aspect, our initialized cluster centroids might leverage some implicit gender- and ethnicity-relevant bias learned from pretrained embeddings, which should be checked when applying our methods into industrial applications.

% \begin{ack}
% Use unnumbered first-level headings for the acknowledgments. All acknowledgments
% More information about this disclosure can be found at: \url{https://neurips.cc/Conferences/2020/PaperInformation/FundingDisclosure}.

% Do {\bf not} include this section in the anonymized submission, only in the final paper. You can use the \texttt{ack} environment provided in the style file to autmoatically hide this section in the anonymized submission.
% \end{ack}

\medskip

\bibliography{neurips_2020}

% \section*{References}
% References follow the acknowledgments. Use unnumbered first-level heading for
% the references. Any choice of citation style is acceptable as long as you are
% consistent. It is permissible to reduce the font size to \verb+small+ (9 point)
% when listing the references.
% {\bf Note that the Reference section does not count towards the eight pages of content that are allowed.}
% \medskip

% \small

% [1] Alexander, J.A.\ \& Mozer, M.C.\ (1995) Template-based algorithms for
% connectionist rule extraction. In G.\ Tesauro, D.S.\ Touretzky and T.K.\ Leen
% (eds.), {\it Advances in Neural Information Processing Systems 7},
% pp.\ 609--616. Cambridge, MA: MIT Press.

\end{document}